\documentclass[conference]{IEEEtran}
\IEEEoverridecommandlockouts
\usepackage{cite}
\usepackage{amsmath,amssymb,amsfonts}
\usepackage{algorithmic}
\usepackage{graphicx}
\usepackage{textcomp}
\usepackage{xcolor}
\usepackage{booktabs}
\usepackage{pifont}
\usepackage{array}
\def\BibTeX{{\rm B\kern-.05em{\sc i\kern-.025em b}\kern-.08em
    T\kern-.1667em\lower.7ex\hbox{E}\kern-.125emX}}
\begin{document}

\title{LCMF: Lightweight Cross-Modality Mambaformer for Embodied Robotics VQA\\
}

\author{\IEEEauthorblockN{1\textsuperscript{st} Zeyi Kang}
\IEEEauthorblockA{\textit{School of Software} \\
\textit{Northwestern Polytechnical University}\\
Xi'an, China \\
kangzeyi@mail.nwpu.edu.cn}
\and
\IEEEauthorblockN{2\textsuperscript{nd} Liang He}
\IEEEauthorblockA{\textit{School of Software} \\
\textit{Northwestern Polytechnical University}\\
Xi'an, China \\
2021050018@nwpu.edu.cn}
\and
\IEEEauthorblockN{3\textsuperscript{rd} Yanxin Zhang}
\IEEEauthorblockA{\textit{School of Software} \\
\textit{Northwestern Polytechnical University}\\
Xi'an, China \\
zhangyanxin2024@mail.nwpu.edu.cn}
\and
\IEEEauthorblockN{4\textsuperscript{th} Zuheng Ming\textsuperscript{*}}
\IEEEauthorblockA{\textit{Laboratoire L2Tl} \\
\textit{University Sorbonne Paris Nord}\\
Paris, France \\
zuheng.ming@univ-paris13.fr}
\and
\IEEEauthorblockN{5\textsuperscript{th} Kaixing Zhao\textsuperscript{*}}
\IEEEauthorblockA{
\textit{School of Software} \\
\textit{Yangtze River Delta Research Institute (Taicang)} \\
\textit{Northwestern Polytechnical University}\\
Xi'an, China \\
kaixing.zhao@nwpu.edu.cn}

\thanks{\textsuperscript{*} Corresponding author.}
}

\maketitle

\begin{abstract}
Multimodal semantic learning plays a critical role in embodied intelligence, especially when robots perceive their surroundings, understand human instructions, and make intelligent decisions. However, the field faces technical challenges such as effective fusion of heterogeneous data and computational efficiency in resource-constrained environments. To address these challenges, this study proposes the lightweight LCMF cascaded attention framework, introducing a multi-level cross-modal parameter sharing mechanism into the Mamba module. By integrating the advantages of Cross-Attention and Selective parameter-sharing State Space Models (SSMs), the framework achieves efficient fusion of heterogeneous modalities and semantic complementary alignment. Experimental results show that LCMF surpasses existing multimodal baselines with an accuracy of 74.29\% in VQA tasks and achieves competitive mid-tier performance within the distribution cluster of Large Language Model Agents (LLM Agents) in EQA video tasks. Its lightweight design achieves a 4.35-fold reduction in FLOPs relative to the average of comparable baselines while using only 166.51M parameters (image-text) and 219M parameters (video-text), providing an efficient solution for Human-Robot Interaction (HRI) applications in resource-constrained scenarios with strong multimodal decision generalization capabilities.
\end{abstract}


\section{Introduction}
In contemporary research, powerful multimodal understanding capabilities have emerged as the foundational element for enabling robotic perception, cognition, and interaction within complex dynamic environments\cite{b1}. In the domain of embodied intelligence, Vision-Language Pre-training (VLP) \cite{c1} has advanced into a critical technological paradigm for the development of sophisticated intelligent robotic systems, offering substantial support for the realization of more intelligent HRI \cite{b6}. Concurrently, to address the challenges inherent in multimodal learning, such as the scarcity of labeled data and the prohibitive costs of annotation, self-supervised learning \cite{c3} has garnered considerable attention and research focus. More precisely, by setting multi-task optimization objectives (multimodal masked modeling, contrastive learning, etc.) \cite{c9}, these methods provide possibilities for common robotic tasks, such as environmental understanding \cite{b2}, decision-making (Visual Question Answering (VQA) \cite{c2}, Embodied Question Answering (EQA) \cite{c12}) or even more advanced cross-modal general understanding \cite{b3}.

However, embodied intelligence \cite{c16} still faces numerous challenges that limit learning capabilities in visual-language decision tasks. At the semantic understanding level, current models \cite{c4,c7,c29} struggle to reconstruct fine-grained mask labels, resulting in an information gap between local features, mask features, and global scene understanding. In addition, the efficiency problem of long sequence modeling cannot be ignored, as the Transformer architecture's computational complexity grows quadratically when processing large-scale sequence data \cite{b4,b7}, making it difficult to achieve optimal trade-offs between cross-modal understanding performance and hardware efficiency.

In response to the above challenges, this paper proposes the lightweight LCMF architecture, which achieves high-quality multimodal understanding and inference acceleration on low-computation robotic platforms. LCMF uses a semantic diffusion mechanism \cite{c4} to address the information gap in multi-scale visual semantics and enhance the ability to model fine-grained masked information. For cross-modal interaction, Cross-Modality Mamba (CMM) extends the Mamba state-space model to the multimodal domain, achieving comprehensive optimization in hardware awareness, time efficiency, and lightweight design. At the level of multimodal semantic fusion, Enhanced Mamba Fusion (EMF) introduces efficient semantic bridging mechanisms and fine-grained feature modulation techniques, enabling the effective integration of heterogeneous modality semantics.

In summary, our contributions include:

1) CMM implements multi-level sharing of state space parameters and parallel modeling of multimodal long sequence semantics, achieving linear computational complexity and inference acceleration.

2) LCMF has implemented a lightweight Mamba-Transformer (Selective SSMs-Attention) architecture in the fields of unimodal feature extraction, multimodal (image, text, video) interaction, and multimodal fusion.

3) Under a significantly reduced parameter scale compared to existing multimodal baselines and LLM Agents, LCMF achieves improved computational efficiency while maintaining strong performance on downstream tasks such as VQA and EQA, demonstrating its effectiveness in efficient multimodal modeling.

The rest of this article is organized as follows.

Section II reviews research on Mamba variant architecture, VQA, EQA.
Section III details the LCMF model architecture, pretraining, fine-tuning, and evaluation methods. 
Section IV describes the experimental setup, performance evaluation, and ablation experiments on specific downstream tasks.
Finally, Section V concludes this article.

\section{RELATED WORK}

\subsection{Multimodal Learning for VQA}

In the field of Visual Question Answering (VQA), multimodal pre-training models have shown significant innovation and breakthroughs\cite{c10}. LXMERT\cite{c17} enhanced the modeling ability of object-word relationships by introducing a dual cross-attention mechanism and three types of pre-training tasks. Unified VLP\cite{c18} innovatively proposed a unified visual-language pre-training framework, which achieves seamless integration of understanding and generation tasks through flexible switching of bidirectional and unidirectional attention. UNITER \cite{c19}, built on a single-stream architecture, enhances visual-language alignment through contrastive learning and region-word alignment tasks, significantly improving VQA performance.

\subsection{Mamba Variant Architecture}

Recently, Mamba \cite{c20} introduced a state-space-based convolutional network architecture that optimizes information flow in deep neural networks by improving convolution operations and skip connection design, further simplifying state transition paths and enhancing network convergence speed and efficiency.
VL-Mamba \cite{c11} proposes a hybrid architectural approach that employs Vision Selective Scan (VSS) mechanism combined with Selective State Space Model layers to optimize cross-modal integration in visual-text processing tasks. Compared to resource-intensive, heavily parameterized Transformer models, the Mamba series provides a more efficient architectural paradigm.

\subsection{Embodied Intelligence in EQA}
Research in the field of EQA\cite{c21} began with Das et al.'s pioneering work, which situated question answering tasks within embodied environments, requiring agents to navigate 3D environments and answer questions. In terms of multimodal fusion, fine-grained vision-language alignment mechanisms such as MCAN\cite{c22} and temporal information integration models like EMQA \cite{b8} have been proposed. Recent research has applied LLM Agents to EQA scenarios, such as EmbodiedGPT\cite{c23}. The NavCoT\cite{c24} model introduces chain-of-thought reasoning in Visual Question Answering (VQA) environments, making the reasoning processes of "what do I see" and "where should I go" explicit. LLM-Planner\cite{c25} breaks down high-level tasks into sequences of sub-goals, forming a complete execution plan in a top-down manner. The SceneParsing\cite{c32} model uses carefully designed scene description prompt templates to guide LLMs in decomposing complex visual scenes into semantic components, and then answers questions based on these components.

\begin{figure*}[t]
    \centering
    \includegraphics[width=0.78\linewidth]{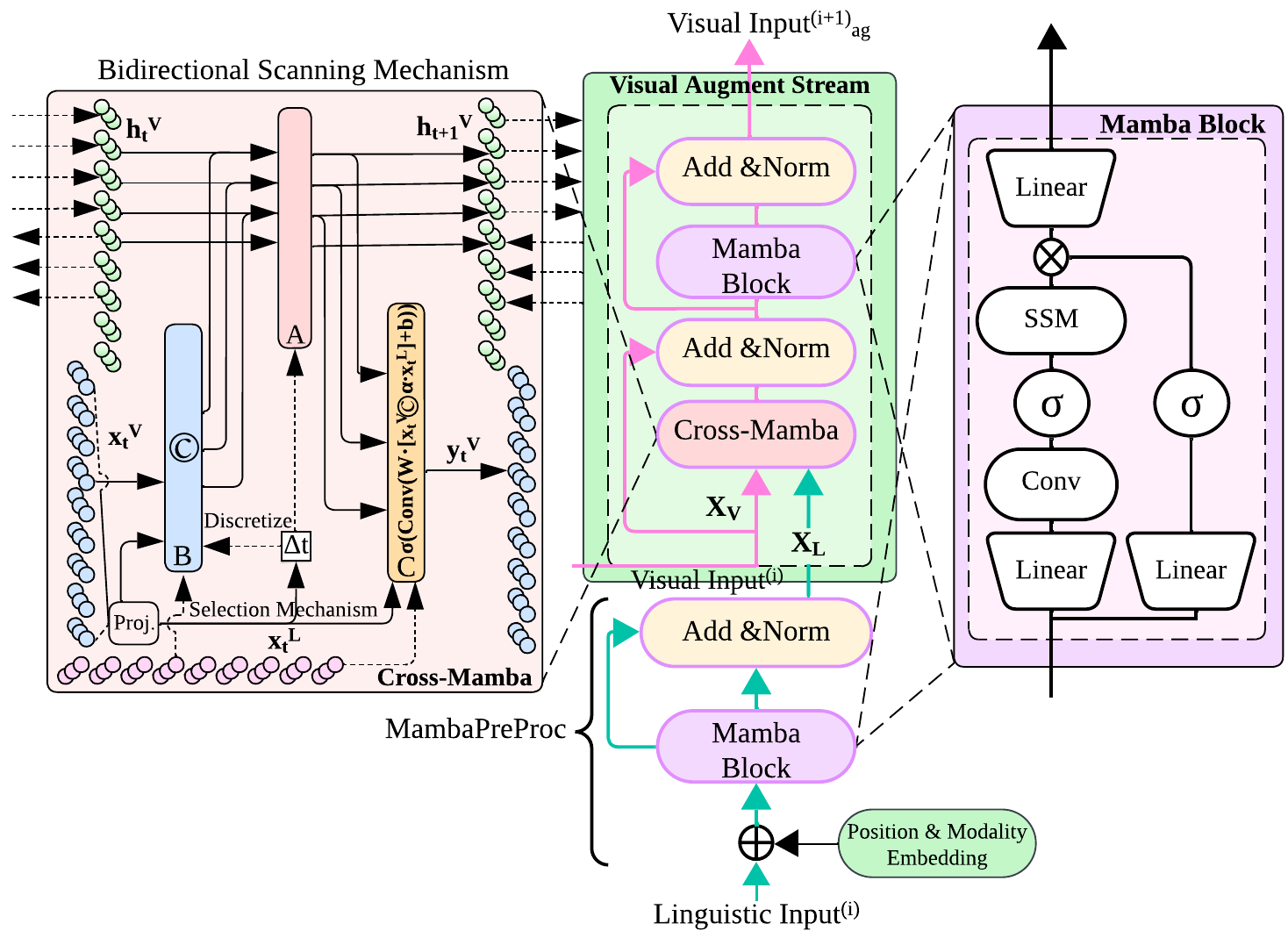}
    \vspace{-8pt}
    \caption{Cross-Modality Mamba Block
(CMM), there are two types, Linguistic Augment Stream \& Visual Augment Stream. Cross-Mamba hierarchically shares the parameters of SSMs (B, C) by concatenating multi-modal features, while maintaining the modal independence of parameters A and $\Delta$.}
    \label{fig:CMM}
    \vspace{-15pt}
\end{figure*}

\section{Methodology}
In this section, we provide a detailed introduction to the design principles of the LCMF model, covering its components, pretraining, fine-tuning, and evaluation strategies.

\subsection{Cross-Modality Mamba}

The proposed Cross-Modality Mamba (CMM) module (see Figure \textcolor{blue}{\ref{fig:CMM}} for details and Figure \textcolor{blue}{\ref{fig:Pretrain}} for its position) achieves deep cross-modal interaction through cross-modal parameter sharing and selective state space transfer, reducing the computational complexity from $O(L^2 \cdot D)$ in traditional Transformer to $O(L \cdot D^2)$. The parallel bidirectional scanning mechanism further reduces inference time from $O(T_V + T_L)$ to $O(\max(T_V, T_L))$.

Cross-Mamba serves as the multimodal interaction component of the CMM module, achieving hierarchical bidirectional parameter sharing between modalities. 
In dynamic SSMs, matrices B and C respectively control information input and output, collaboratively capturing long-range dependencies within sequences. The cross-modal parameter sharing mechanism enables coordinated selective decision-making across multiple modalities.
Each modality is preprocessed through MambaPreProc to obtain $X_V$ and $X_L$ via selective SSMs. Taking vision as an example:


\subsubsection{\textbf{Feature Projection And Expansion}}
  \begin{itemize}
  \item $[U_{\text{SSM}}^{(V)}, U_{\text{conv}}^{(V)}] = Split(Linear_{\text{proj}}^{(V)}(LayerNorm(X_V)))$
  \end{itemize}

where: $U_{\text{SSM}}^{(V)}$ captures long-range dependencies and sequential information; $U_{\text{conv}}^{(V)}$ captures local features and short-range dependencies; $Split(\cdot)$ divides the projected features into SSM and convolution branches.


To balance modal specificity and cross-modal interaction, we adopt a layer-wise differentiated interaction strategy.
  \begin{itemize}
  \setlength{\itemsep}{0.5em}
  \item $A^{(V)}=\sigma(W_A \cdot U_{\text{SSM}}^{(V)} + b_A)$ 
  \item $B^{(V)} = SiLU(Conv1D(W_{B} \cdot [U_{\text{SSM}}^{(V)}; \alpha_l \cdot U_{\text{SSM}}^{(L)}] + b_{B}))$
  \item $C^{(V)} = SiLU(Conv1D(W_{C} \cdot [U_{\text{SSM}}^{(V)}; \alpha_l \cdot U_{\text{SSM}}^{(L)}] + b_{C}))$
  \item $\Delta^{(V)} = softplus(W_{\Delta V} \cdot U_{\text{SSM}}^{(V)} + b_{\Delta V})$
  \end{itemize}
  
where: $l$ is the current layer index, $L$ is total layer count, $\alpha_l = l/L$ controls interaction strength (weaker in shallow, stronger in deep layers). $A^{(V)}$ is the SSM state transition matrix, $B^{(V)}$ and $C^{(V)}$ are SSM input and output projection matrices. $\Delta^{(V)}$ is the time-step parameter. $softplus(\cdot)$ ensures $\Delta$ is positive.

\subsubsection{\textbf{SSM Param Discretization \& Seq Process}}
  \begin{itemize}
  \setlength{\itemsep}{0.5em}
  \item $\bar{A}^{(V)} = e^{A^{(V)} \cdot \Delta^{(V)}}$, $\bar{B}^{(V)} = \frac{e^{A^{(V)} \cdot \Delta^{(V)}} - 1}{A^{(V)}} \cdot B^{(V)}$
  \item $V_{\text{SSM}}^{(V)} = SelectiveScan(U_{\text{SSM}}^{(V)}, \bar{A}^{(V)}, \bar{B}^{(V)}, C^{(V)})$
  \end{itemize}
  
where: $\bar{A}^{(V)}$ and $\bar{B}^{(V)}$ are the discretized SSM parameters, $SelectiveScan(\cdot)$ performs the sequential state space computation with hardware-optimized scanning.

\subsubsection{\textbf{Parallel Convolution and Output}}
  \begin{itemize}
  \setlength{\itemsep}{0.5em}
  \item $V^{(V)} = V_{\text{SSM}}^{(V)} \odot \sigma(Conv1D(U_{\text{conv}}^{(V)}))$
  \item $X'_V = LayerNorm(X_V) + Linear_{\text{out}}^{(V)}(V^{(V)})$
  \item $Z^{(V)} = LayerNorm(X'_V)$
  \end{itemize}

where: $V^{(V)}$ is the gated output combining SSM and convolution branches, $\odot$ denotes element-wise multiplication.

\begin{figure*}
    \centering
    \includegraphics[width=1.0\linewidth]{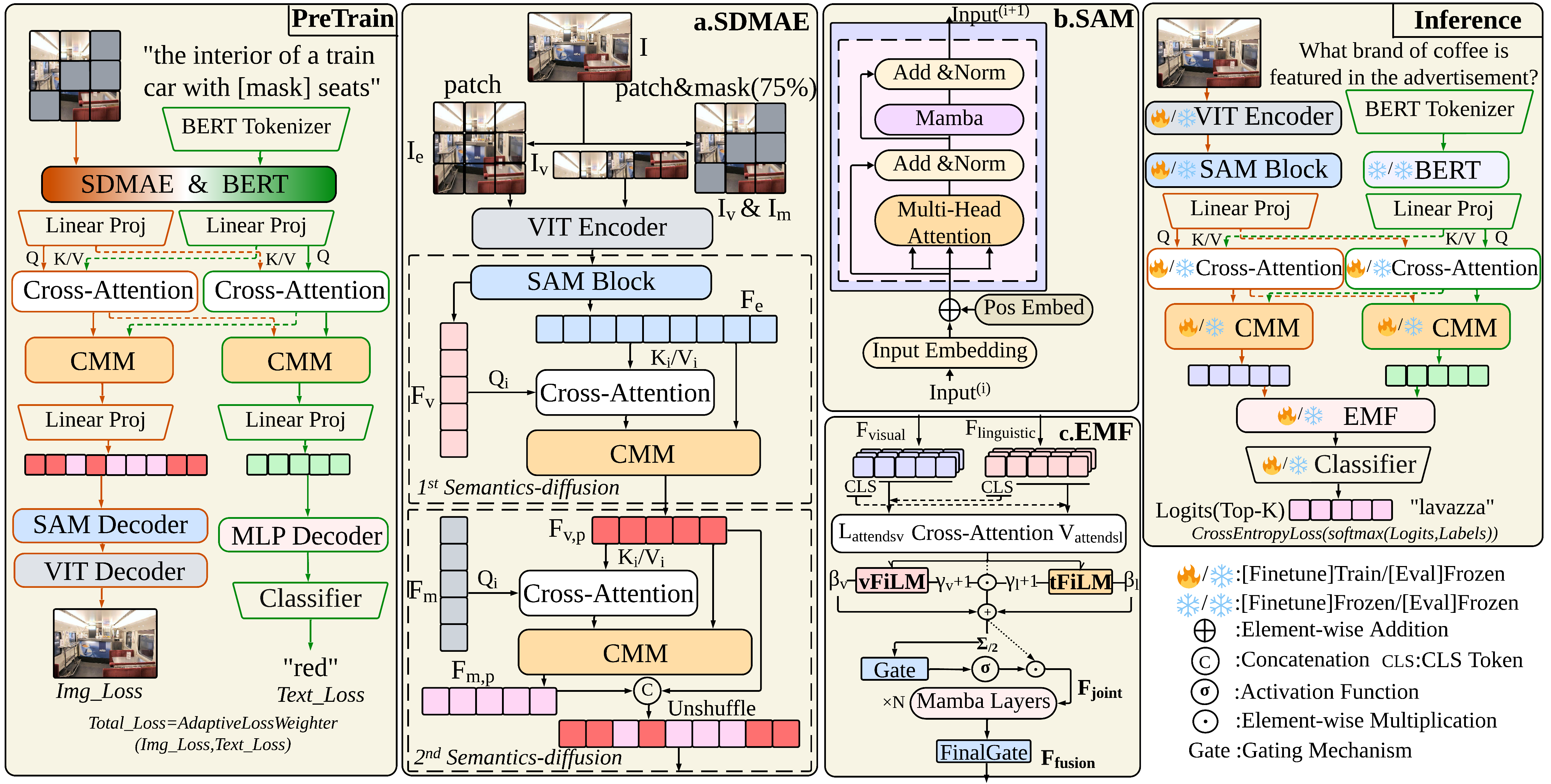}
    \vspace{-20pt}
    \caption{The LCMF PreTrain task framework and downstream task reasoning framework. Subgraph a shows the details of visual semantic multi-level diffusion in SDMAE; subgraph b displays the SAM unimodal SelfAttention-Mamba cascading architecture; subgraph c illustrates the details of the EMF.}
    \label{fig:Pretrain}
    \vspace{-15pt}
\end{figure*}

CMM is a sequence processing mechanism based on selective parameter-sharing SSMs that models the dynamic changes of sequence data through a set of multimodal hidden states.
Hardware-specific implementation of block scanning further optimizes GPU parallelism\cite{c20}:
\begin{equation}
h_{b,j} = \bar{A}_{b,j} h_{b,j-1} + \bar{B}_{b,j} x_{b,j}
\end{equation}
\begin{equation}
y_{b,j} = C_{b,j} h_{b,j} + D x_{b,j} 
\end{equation}


where $h_{b,j} \in \mathbb{R}^{N}$ and $x_{b,j} \in \mathbb{R}^{D}$ are the hidden state and input vectors at position $j$ in block $b$, respectively. $\bar{A}_{b,j} \in \mathbb{R}^{N \times N}$ and $\bar{B}_{b,j} \in \mathbb{R}^{N \times D}$ are the discretized state transition and input projection matrices. $y_{b,j} \in \mathbb{R}^{D}$ is the output vector, with $C_{b,j} \in \mathbb{R}^{D \times N}$ as the state-to-output projection matrix and $D \in \mathbb{R}^{D \times D}$ as the skip connection matrix. Here, $b \in \{1,2,\ldots,B\}$ and $j \in \{1,2,\ldots,L_b\}$ denote the block index and sequence position, where $N$, $D$, $B$, and $L_b$ represent hidden state dimension, input/output feature dimension, number of parallel blocks, and block sequence length, respectively. This architecture forms the foundation of our CMM approach for multi-modal contexts.

\subsection{Semantics-diffusion Masked Autoencoder} 

The SDMAE module (as illustrated in Figure \textcolor{blue}{\ref{fig:Pretrain}} subgraph a) consists of three core elements: a patchify module, a visual feature extraction module, and a two-stage cross-semantic diffusion module.
The visual feature extraction module consists of stacked layers of ViT and Self-Attention Mamba (SAM) blocks, denoted as $E$. As shown in Figure \textcolor{blue}{\ref{fig:Pretrain}} subgraph b, SAM combines the global receptive field of self-attention with the efficient sequential processing capability of selective SSMs, thereby optimizing visual feature extraction.

The image \( I \)  is partitioned into patches to form a full patch set $I_e$. After the random masking operation, $I_e$ is divided into two disjoint subsets: the set of visible patches \(I_{v}\) and the set of masked patches \(I_{m}\). $E$ transforms the image patch sets $I_e$ and $I_v$ into feature representations $F_e$ and $F_v$, respectively.

The two-stage cross-semantic diffusion module achieves visual semantic enhancement through a hybrid architecture of Cross-Attention \cite{b4} and CMM. Inspired by CAE \cite{b9} and MAE-MCAD \cite{c29} , SDMAE adopts an explicit context modeling mechanism, implementing a two-stage diffusion process from global to local visual semantics, then to mask semantics.
First stage: $F_e$ and $F_v$ undergo semantic enhancement to obtain $F_{v,p}$; Second stage: $F_{v,p}$ and $F_m$ are processed to obtain $F_{m,p}$. Both are concatenated and reordered to form visual semantics for image reconstruction pretraining. 

The two-stage diffusion process is the same, taking the first stage as an example. In the Cross Attention layer, visible features $F_v \in \mathbb{R}^{N_v \times d}$ serve as queries, while complete features $F_e \in \mathbb{R}^{N_e \times d}$ serve as key-values:

\begin{equation}
F_v' = Softmax \left( \frac{QF_v(KF_e)^T}{\sqrt{d_k}} \right) \cdot VF_e
\end{equation}

$F_v'$ and $F_e$ are further processed through the CMM layer to enhance visual semantics, yielding $F_{v,p}$.

\subsection{Training Strategy}
The overall training strategy is shown in Figure \textcolor{blue}{\ref{fig:Pretrain}}. Based on the self-supervised learning task paradigm, we use multimodal masked reconstruction as the pretraining optimization objective. The model extracts multimodal features through the SDMAE-Bert encoder, uses a cascaded CMM-CrossAttention architecture for cross-modal interaction, and finally completes the reconstruction tasks through corresponding decoders (ViT-SAM decoder and MLP decoder).

\textbf{Adaptive Multi-Loss Weighter:}
To balance the loss weights of multimodal tasks, we designed an adaptive regulator that tracks loss trends using an Exponential Moving Average (EMA) \cite{c31} and dynamically adjusts weights based on relative changes between tasks. To prevent drastic fluctuations, the adjustment magnitude is constrained, ensuring smooth and stable weight updates during training.

The overall loss function is composed of image reconstruction loss and masked language modeling loss, aiming to jointly optimize visual and textual representations. The task weights are dynamically optimized through an Adaptive Multi-Loss Weighter.

\subsection{Downstream Task Settings and Frameworks}
As shown in Figure \textcolor{blue}{\ref{fig:Pretrain}}, We finetune the pretrained parameters of the multimodal encoder on VQA/EQA downstream tasks, transferring the general multimodal encoding capabilities to specific Human-Robot Interaction scenarios.

\textbf{Enhanced Mamba Fusion:}
As shown in Figure \textcolor{blue}{\ref{fig:Pretrain}} subgraph c, EMF framework combines the advantages of selective SSMs and cross-attention in fusing multimodal features. Visual-language features are adjusted and embedded into a joint space with CLS Token extraction, using mirrored operations between modalities (taking vision as an example), with the fusion process as follows:
\begin{equation}
V_{attends_L} = CrossAttention(Q=V_{cls}, K\&V=L_{cls})
\end{equation}

Our key innovation is the Feature Linear Modulation (FiLM) applied to cross-attention outputs. By generating scaling and offset factors through linear transformation and applying them to attention representations, we achieve adaptive feature modulation, enabling features from one modality to dynamically adjust the representation of another modality.
\begin{equation}
[\gamma_v, \beta_v] = W_{film_v} \cdot V_{attends_L}
\end{equation}
\begin{equation}
V_{modulated} = (1 + \gamma_v) \odot L_{attends_V} + \beta_v
\end{equation}

Next, we aggregate modulated features and apply a gating mechanism:
\begin{equation}
F_{joint} = (V_{modulated} + L_{modulated})/2
\end{equation}
\begin{equation}
F_{gated} = F_{joint} \odot \sigma(W_g \cdot F_{joint})
\end{equation}

Finally, Mamba layers process the fused features to generate the multimodal representation, where $X_0 = F_{gated}, i \in \{1,2,...,N\}$:
\begin{equation}
X_i = LayerNorm(X_{i-1} +Mamba(X_{i-1}))
\end{equation} 
\begin{equation}
F_{output} = LayerNorm(X_N \odot \sigma(W_f \cdot X_N))
\end{equation}

The EMF framework optimizes cross-modal attention via FiLM, enhances task-specific features through adaptive gating, and captures cross-modal dependencies using multi-level Mamba fusion, forming a lightweight and efficient multimodal fusion architecture.

\section{EXPERIMENTS}
The experimental part covers various aspects such as multimodal reconstruction pre-training, downstream tasks VQA/EQA, efficiency improvement, parameter comparison, and actual deployment on robots.

\subsection{Experimental Setup}

\begin{table}[htbp!]
\centering
\caption{LCMF Configuration}
\label{tab:table1}
\begin{tabular}{l|ccc}
\toprule
Configuration & Pretrain & VQA & EQA \\
\midrule
Hidden Dimensions & \multicolumn{3}{c}{768} \\
FiLM Output Dimensions & - & 1536 & 1536 \\
FiLM Modulation Dims ($\gamma$, $\beta$) & - & 768 & 768 \\
Optimizer & \multicolumn{3}{c}{AdamW} \\
Base Learning Rate  & 5e-5 & 5e-5 & 1e-4 \\
Batchsize  & 128 & 164 & 4 \\
Epoch  & 300 & 300 & 400 \\
Learning Rate Schedule  & 1Cycle & 1Cycle & CosAnn \\
SDMAE Layers & - & 4 & 4 \\
Loss Weighter EMA \cite{c31} & 0.95 & - & - \\
Answer Vocab \cite{c2} & - & 1000 & 366 \\
\midrule
Bert & \multicolumn{3}{c}{bert-base-uncased} \\
Text Mask Prob \cite{c5} & \multicolumn{3}{c}{15\%} \\
Masking Strategy \cite{c5} & \multicolumn{3}{c}{80/10/10 [M/R/O]} \\
\midrule
Image Resolution & \multicolumn{3}{c}{$224^2$ pixels} \\
Image Mask Prob \cite{c7} & \multicolumn{3}{c}{75\%} \\
Mask Token Sample & \multicolumn{3}{c}{Random sampling} \\
Norm Mean \& std & \multicolumn{3}{c}{\ding{51}}  \\
\midrule
Video Sample Frames & \multicolumn{3}{c}{50} \\
Sampling Method & \multicolumn{3}{c}{stratified} \\
\bottomrule
\end{tabular}
\end{table}

The model hyperparameters and data preprocessing configurations are detailed in Table \textcolor{red}{\ref{tab:table1}}. These configurations are determined after extensive ablation studies. In the downstream tasks of VQA and EQA, we use Top-K sampling to construct a closed vocabulary of high-frequency answers, aiming to achieve an optimal balance between vocabulary size and coverage.

\textbf{Metrics:} In the pretraining phase, we use image mean square error (pixel-level MSE), text cross-entropy loss (CE Loss) to comprehensively measure the model convergence level. For VQA and EQA downstream tasks, we treat them as multi-classification problems and report a series of key metrics: overall accuracy, accuracy by question type, model parameter count, percentage reduction in computational effort (FLOPs), and inference speed.

\subsection{Multimodal Reconstruction}

\begin{figure}[htbp]
    \centering
    \includegraphics[width=0.9\columnwidth]{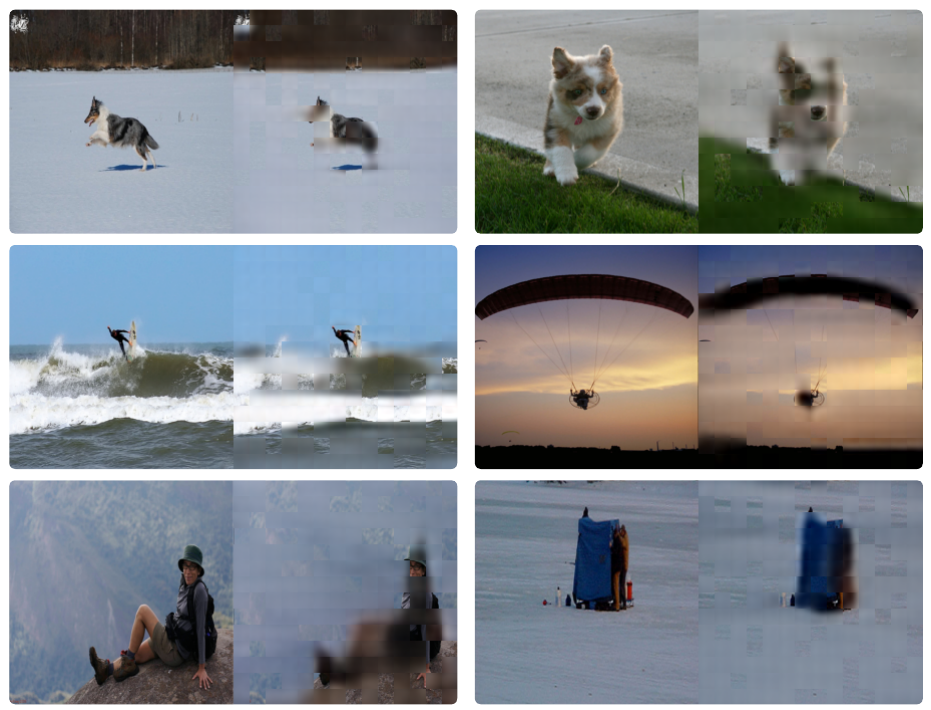}
    \caption{\textbf{Image Reconstruction Example}}
    \label{fig:image4}
\end{figure}

During the pre-training phase, we use the Flickr8k\cite{c27} dataset for multimodal masked reconstruction. To validate whether semantic information of images was successfully acquired, as shown in Figure \textcolor{blue}{\ref{fig:image4}}, we compared the results of image mask modeling (right) with the display of original images (left), and observed that the image semantics are credible\cite{c7}. Pixel-level MSE has optimization limitations, and reconstruction is not the core objective of the model. We focus more on pretraining multimodal semantic representations and transferring their generalizable semantic understanding to downstream tasks such as VQA (EQA).

\begin{figure}[htbp]
    \centering
    \includegraphics[width=0.7\columnwidth]{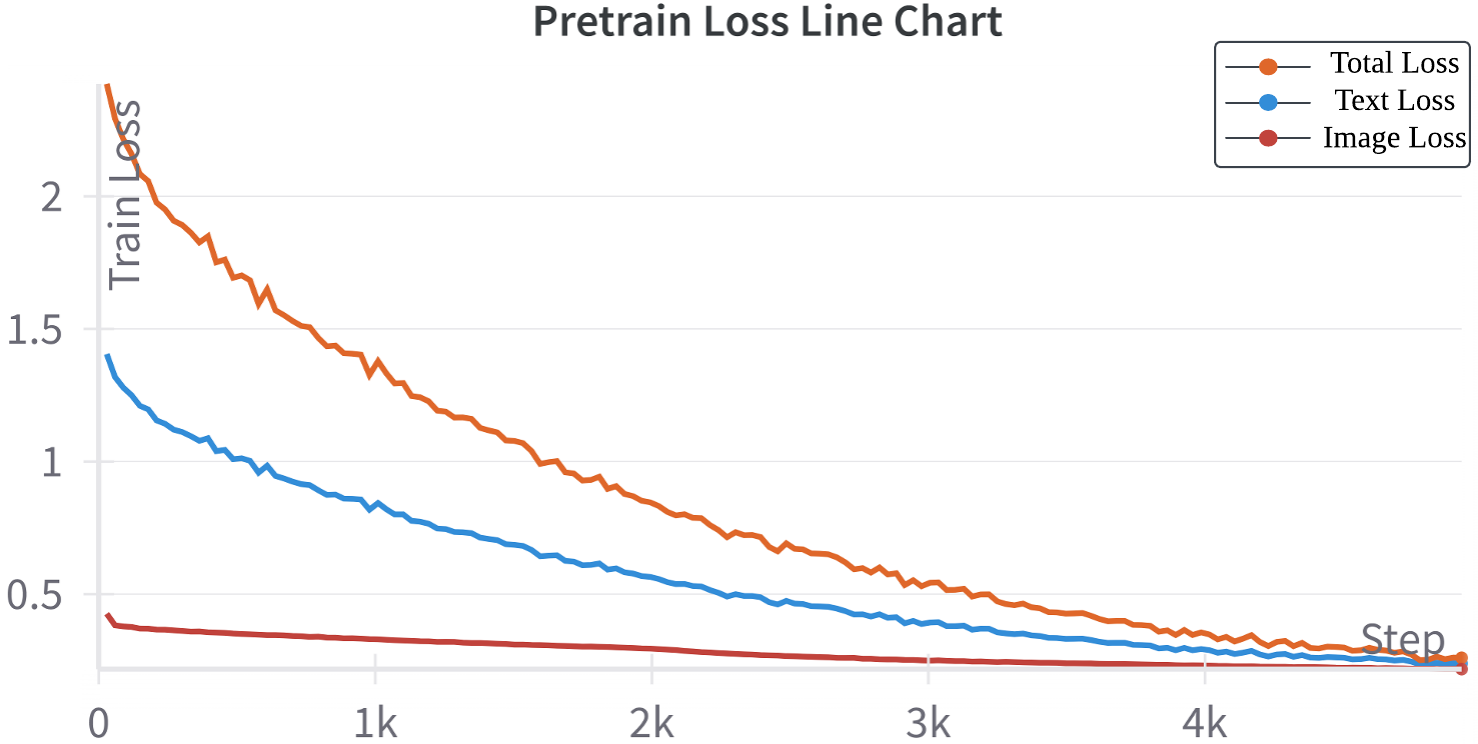}
    \caption{\textbf{Reconstruction Loss Line Chart}}
    \label{fig:image5}
\end{figure}

\begin{figure}[htbp]
    \centering
    \includegraphics[width=0.7\columnwidth]{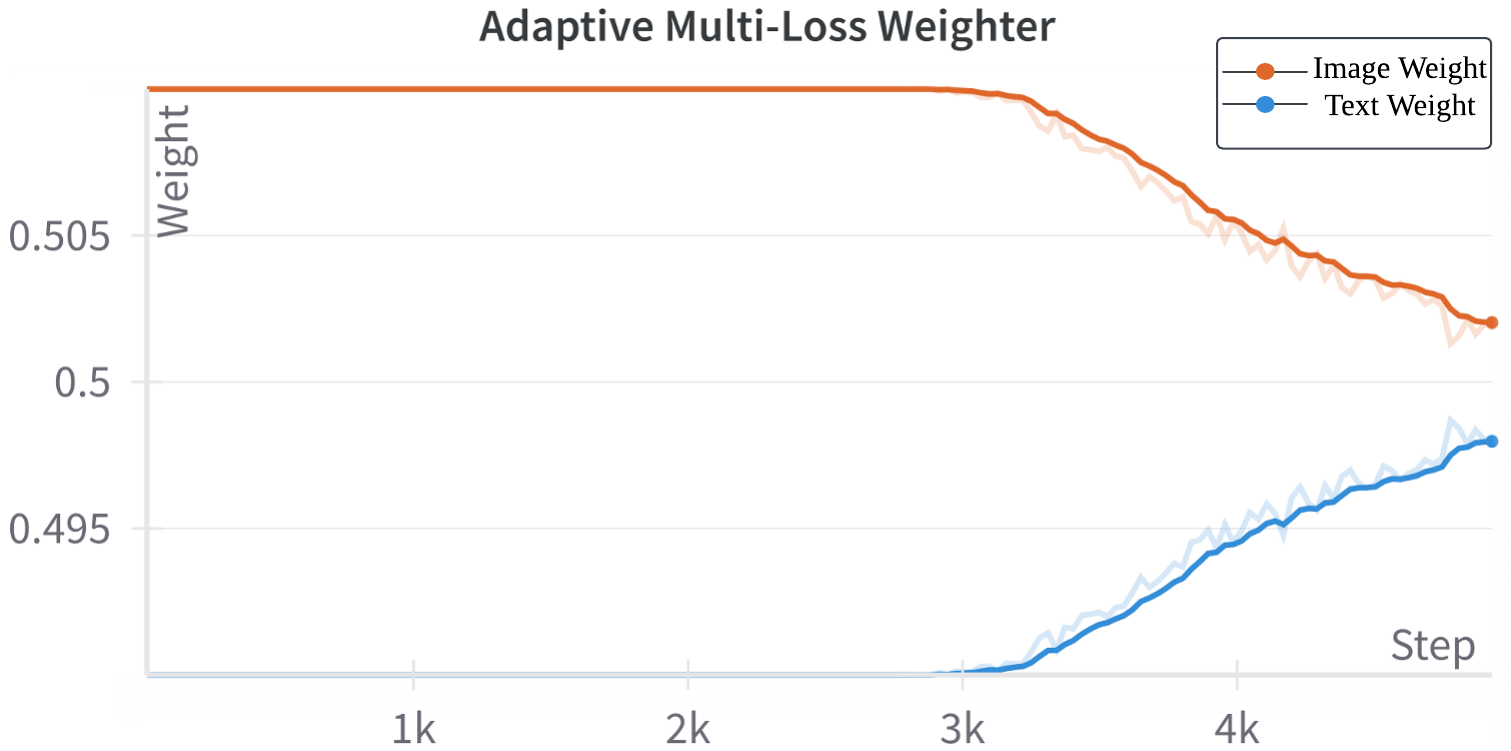}
    \caption{\textbf{Adaptive Multi-Loss Weighter Line Chart}}
    \label{fig:image6}
\end{figure}

The training results are shown in the Figure \textcolor{blue}{\ref{fig:image5}}, the differences in the magnitude of modal losses indicate that the scale of information entropy is fundamentally different. All loss components consistently converge during the training process, and the gain begins to decrease after 300 epochs. As shown in Figure \textcolor{blue}{\ref{fig:image6}}, Adaptive Multi-Loss Weighter smooths the fluctuations of modal losses, preventing short-term fluctuations from excessively affecting weight distribution. The weight curve remains stable in the early stages and gradually converges in the later stages. This effectively alleviates amplitude differences between modalities, avoids the dominance of a single modality, and achieves optimal synergy.

\subsection{Visual Question Answer(VQA)}

\textbf{VQA Performance:} As shown in Table \textcolor{red}{\ref{tab:table3}, \ref{tab:table4}}, LCMF achieved an accuracy of 74.29\% on the VQAv2\cite{c2} validation set. We calculated the Mean-average Accuracy (mAA) and achieved a score of 70.9\%. LCMF performed best in the "Yes/No" and "Number" categories, achieving 90.6\% and 59.4\% accuracy, respectively. These results show that the cross-modal fusion mechanism of LCMF can effectively capture the complex associations between images and texts, providing more comprehensive and in-depth decision-making capabilities for visual question answering tasks.

\begin{table}[h]
  \centering
  \caption{MODEL PERFORMANCE COMPARISON ON THE VQA TASK}
  \label{tab:table3}
  \begin{tabular}{lcccc}
    \toprule  
    \textbf{Model} & \textbf{Accuracy} & \textbf{Yes/No} & \textbf{Number} & \textbf{Other} \\
    \midrule  
    \textbf{ViLBERT \cite{c13}} & 70.9\% & - & - & - \\
    \textbf{LXMERT \cite{c17}} & 72.5\% & 88.2\% & 54.2\% & \textbf{63.1\%} \\
    \textbf{Unified VLP \cite{c18}} & 70.7\% & 87.4\% & 52.1\% & 60.5\% \\
    \textbf{UNITER \cite{c19}} & 73.4\% & - & - & - \\
    \textbf{LCMF} & \textbf{74.29\%} & \textbf{90.6\%} & \textbf{59.4\%} & 62.9\% \\
    \bottomrule  
  \end{tabular}
\end{table}

\begin{table}[h]
  \centering
  \caption{Model Efficiency and Performance Metrics}
  \label{tab:table4}
  \begin{tabular}{lccc}
    \toprule  
    \textbf{Model} & \textbf{mAA} & \textbf{Param} & \textbf{FLOPs}($\times 10^9$) \\
    \midrule  
    \textbf{ViLBERT \cite{c13}} & - & 223.68M  & 45.7\\
    \textbf{LXMERT \cite{c17}} & 68.5\% & 207.94M & 9.6410 \\
    \textbf{Unified VLP \cite{c18}} & 66.7\% & 167.80M & 47.84 \\
    \textbf{UNITER \cite{c19}} & - & 303.48M & 61.2 \\
    \textbf{LCMF(Inference)} & \textbf{70.9\%} & \textbf{166.51M} & \textbf{9.4512} \\
    \bottomrule  
  \end{tabular}
\end{table}

\textbf{Parameter and Computational Efficiency Analysis:} As shown in Table \textcolor{red}{\ref{tab:table4}}, 
LCMF model achieves optimal performance with only 166.51M parameters and 9.4512$\times$10$^9$ FLOPs computational complexity. This result demonstrates that our cascade attention and lightweight cross-modal fusion mechanism design based on CMM is highly efficient. With the support of the Cross-Mamba structure, modal parameter sharing is implemented and more accurate visual-language correspondences are captured with linear complexity, greatly relieving the computational pressure of the Attention mechanism. LCMF achieves inference acceleration while reducing redundant parameters.

\begin{figure}[htbp]
    \centering
    \includegraphics[width=\columnwidth]{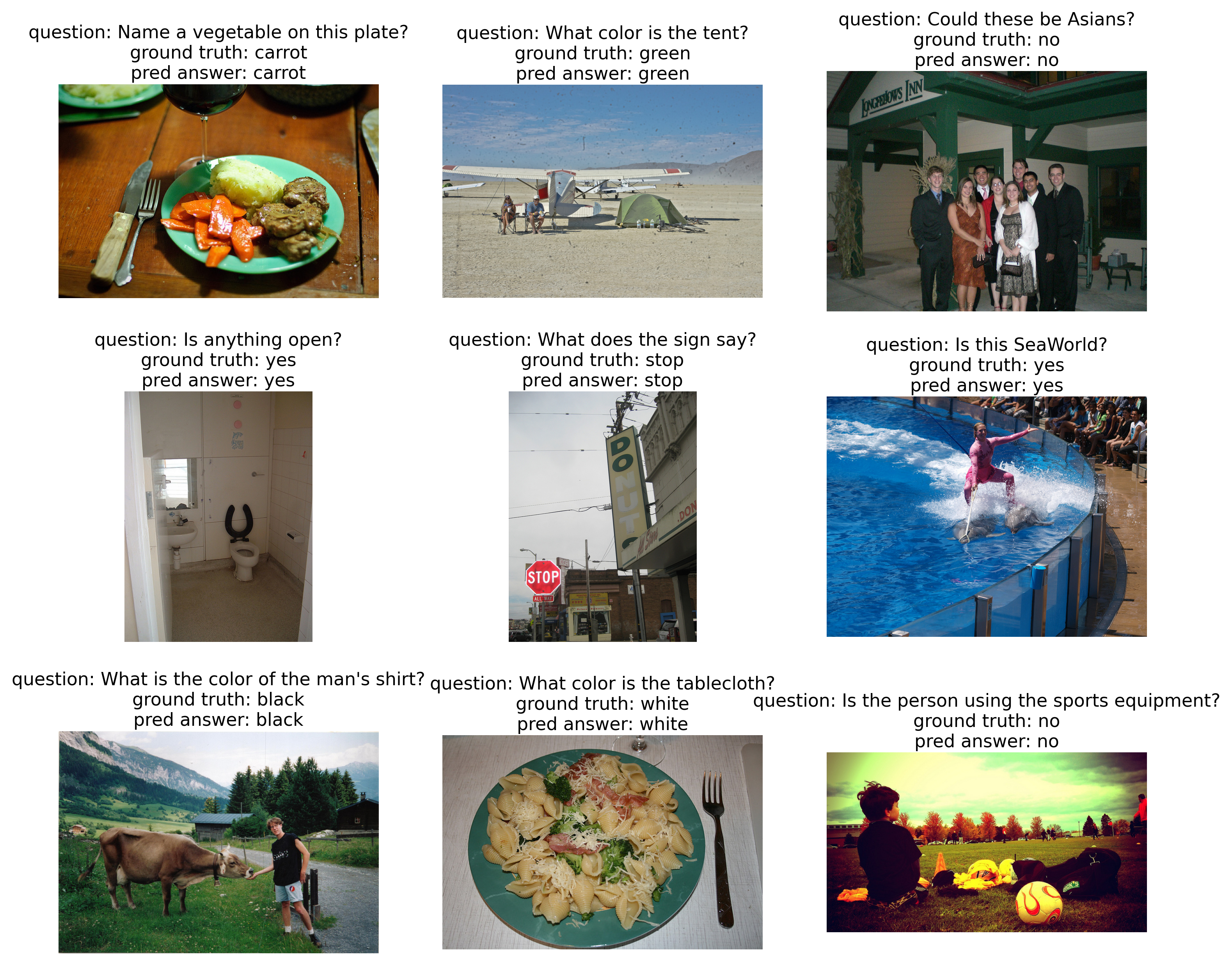}
    \caption{\textbf{VQA Task Visualization}}
    \label{fig:image7}
\end{figure}

Finally, as shown in the Figure \textcolor{blue}{\ref{fig:image7}}, LCMF successfully achieves accurate predictions based on visual and question clues, demonstrating its decision-making capability enhanced through multimodal knowledge integration.

\subsection{Ablation Study}

To demonstrate the contributions of different components in the LCMF framework, we conducted an ablation study based on the controlled variable method. 
As shown in Table \textcolor{red}{\ref{tab:table6}}, this analysis investigates the impact of removing or replacing key modules (including Cross-Attention, CMM Blocks, and SAM Blocks) on downstream VQA task performance. 

\begin{table}[h]
  \centering
  \caption{Ablation Study Results On VQA}
  \label{tab:table6}
  \begin{tabular}{lccc}
    \toprule  
    \textbf{Model} & \textbf{Param}  & \textbf{VQA Acc} & \textbf{FLOPs}($\times 10^9$)\\
    \midrule  
    \textbf{LCMF(Inference)} & 166.51M & \textbf{74.29\%} & 9.4512 \\
    \textbf{w/o Cross-Attention} & 161.78M & 69.21\% & 8.572 \\
    \textbf{w/o CMM} & \textbf{152.25M} & 65.27\% & \textbf{8.087} \\
    \textbf{w/o SAM} & 161.74M & 67.58\% & 9.0312 \\
    \bottomrule  
  \end{tabular}
\end{table}

More precisely, LCMF enhances cross-modal information extraction by maximizing conditional mutual information. After removing Cross-Attention, the number of parameters is reduced to 161.78M (a decrease of about 2.8\%), and the computational cost is reduced to 8.572$\times$10$^9$ FLOPs, but the accuracy is reduced to 69.21\%. This verifies the importance of cross-attention in reducing conditional entropy.

Removing the entire CMM module resulted in a significant accuracy drop to 65.27\%, a decrease (9.02\%) that far exceeds the impact of removing only Cross-Attention (5.08\%). This indicates that the parameter sharing strategy of the CMM module, combined with the linear state space model, effectively models multimodal semantic relationships and reduces computational burden.

After removing SAM, the model accuracy dropped to 67.58\%, which proves that SAM effectively combines the advantages of selective SSMs and the self-attention mechanism. It significantly improves the processing efficiency of long sequence information and more accurately captures important feature associations in single-modal data.

\subsection{Robot deployment: EQA}
\begin{figure}[htbp]
    \centering
    \includegraphics[width=0.9\columnwidth]{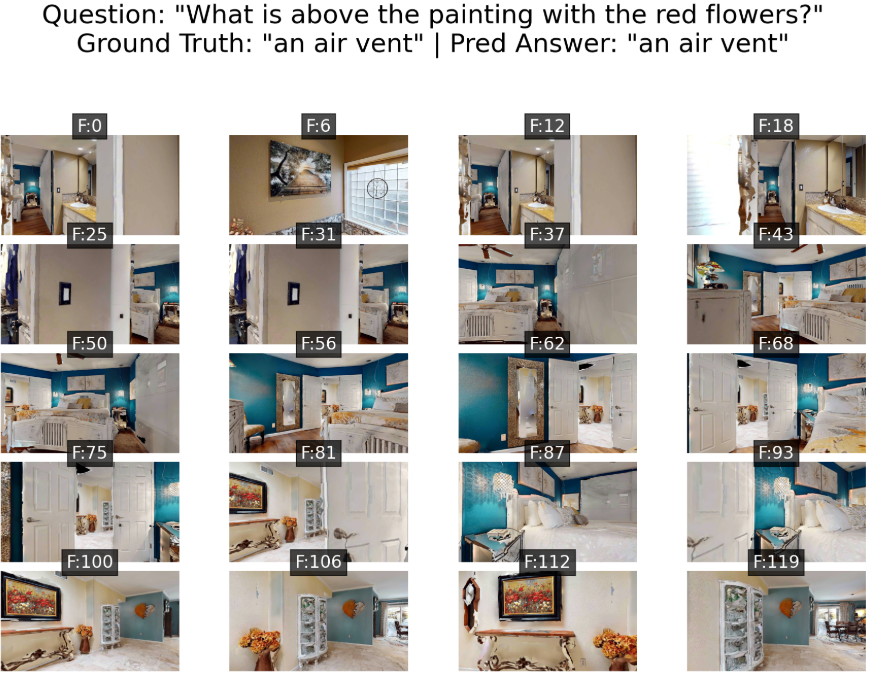}
    \caption{\textbf{OpenEQA Task Visualization}}
    \label{fig:image8}
\end{figure}

In the OpenEQA Habitat-Matterport 3D\cite{c12} benchmark, we extend the visual extraction capabilities of LCMF to the field of video processing\cite{c26}. Through the temporal component, we capture and analyze the dynamic visual cue associations between video frame sequences. As shown in Figure \textcolor{blue}{\ref{fig:image8}}, by sampling key video cues, LCMF can achieve high-quality multimodal reasoning and decision-making in complex human-robot interactive question-answering scenarios, and build the cognitive and decision-making capabilities of humanoid robots in the real world.


We propose two evaluation methods\cite{b5}: one is to train the entire video set and use different video frames for verification sampling (non-Zero Shot); the other is to divide the video set into non-overlapping subsets and perform training and evaluation separately (Zero Shot).

\begin{figure}[htbp]
    \centering
    \includegraphics[width=\columnwidth]{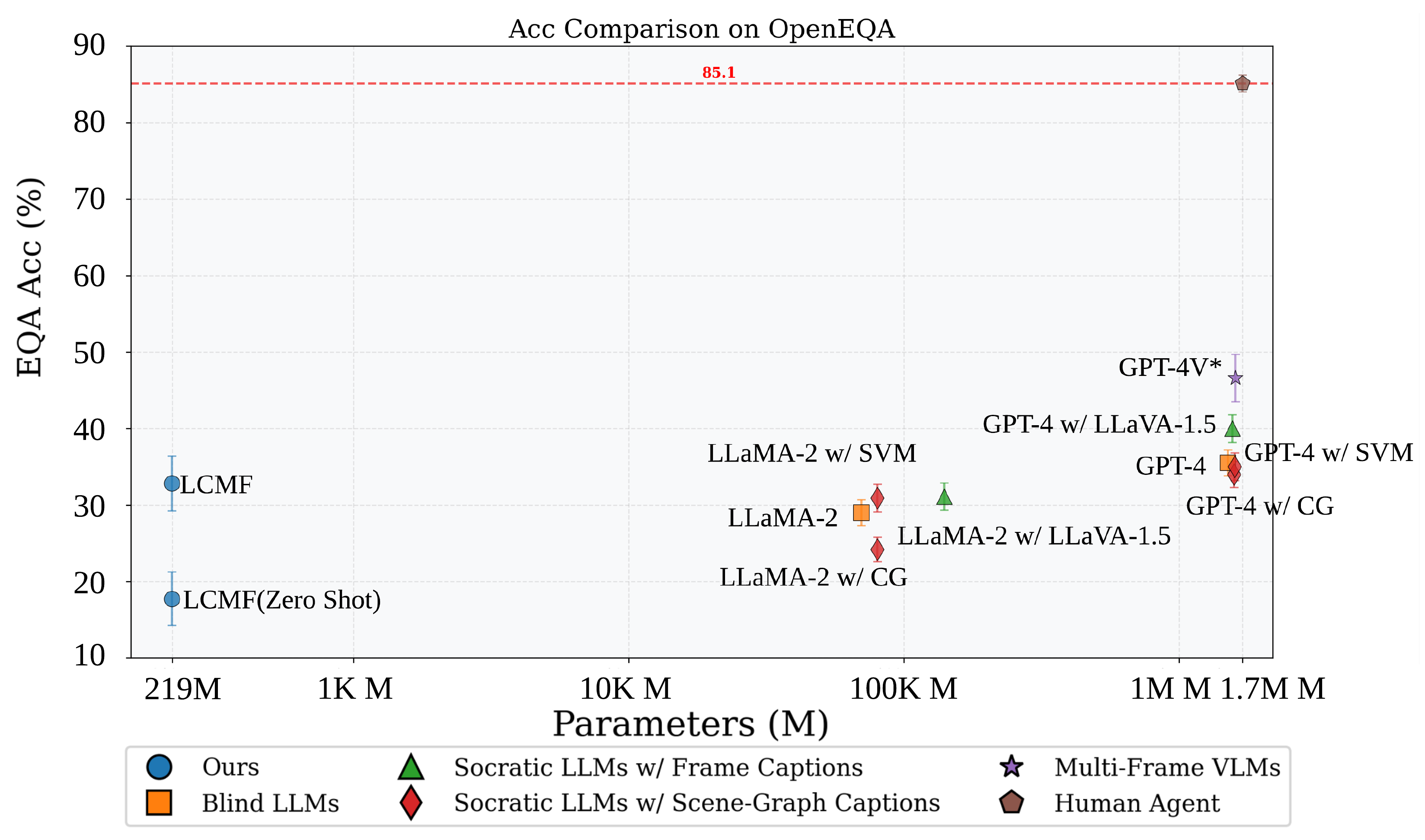}
    \caption{\textbf{Accuracy Comparison On OpenEQA}}
    \label{fig:image1}
\end{figure}

As shown in Figure \textcolor{blue}{\ref{fig:image1}}, LCMF is compared with various mainstream LLM Agents reported in the OpenEQA benchmark \cite{c12}, including Blind LLMs, Socratic LLMs with Frame Captions, Socratic LLMs with Scene-Graph Captions, and Multi-Frame Visual-Language Models (VLMs). Benefiting from their large-scale parameters, LLM Agents are able to model complex patterns and relationships, which strongly supports their general-purpose intelligence across diverse tasks. Their accuracy on the OpenEQA dataset ranges from 21\% to 46\%. The best-performing model is GPT-4V* released by OpenAI (46.6\%±3.1). LCMF (non-Zero Shot) achieves an accuracy of 32.82\%±3.6, while the Zero Shot version reaches 17.74\%±3.5, which is generally weaker compared to the performance of LLM Agents. The parameter scale of LCMF differs significantly from that of LLM Agents, as LCMF is specifically designed to build intelligent decision-making capabilities for the limited scope of the OpenEQA benchmark, whereas LLM Agents possess the general ability to process world knowledge.

\section{DISCUSSION}
In this study, our proposed LCMF was comprehensively compared with existing baseline solutions across image-text and video-text dimensions. The experimental results demonstrate that LCMF achieves competitive performance while significantly reducing model complexity, validating its effectiveness and superiority in cross-modal understanding tasks. Although this paper emphasizes the lightweight design philosophy of the model, we acknowledge the lack of in-depth quantitative analysis of key performance indicators, including model inference latency, memory consumption, and energy consumption, which are core considerations in practical deployment. In future work, we will further refine the model deployment performance evaluation system to provide more comprehensive and reliable performance benchmarks for the practical application of LCMF in edge computing scenarios.

\section{CONCLUSION}
This paper proposes LCMF, a lightweight Mamba-Transformer (Selective SSMs-Attention) architecture, which builds strong general multimodal capabilities for downstream VQA/EQA tasks. It provides a theoretical foundation for lightweight multimodal understanding and intelligent decision-making in robotic systems. 

\section{ACKNOWLEDGMENTS}
The authors gratefully acknowledge the financial supports by different fundings. Kaixing Zhao is supported by the National Natural Science Foundation of China (No. 62407035 ), the China Postdoctoral Science Foundation (No. 2024M754226) and General Project of Taicang Basic Research Plan (No. TC2022JC11). Authors also thank the support of Hyper Creative Industry (Suzhou) Technology Co., Ltd.


\bibliographystyle{IEEEtran}
\bibliography{main}

\end{document}